\documentclass[sigconf]{acmart}
\AtBeginDocument{%
  \providecommand\BibTeX{{%
    \normalfont B\kern-0.5em{\scshape i\kern-0.25em b}\kern-0.8em\TeX}}}

\setcopyright{acmlicensed}
\copyrightyear{2024}
\acmYear{2024}
\acmDOI{XXXXXXX.XXXXXXX}

\acmISBN{978-1-4503-XXXX-X/18/06}

\begin{document}

\title[CC-GPX: Extracting High-Quality Annotated Geospatial Data from Common Crawl]{CC-GPX: Extracting High-Quality Annotated Geospatial Data from Common Crawl}

\author{Ilya Ilyankou}
\authornote{Corresponding author.}
\email{ilya.ilyankou.23@ucl.ac.uk}
\orcid{0009-0008-7082-7122}

\author{Meihui Wang}
\email{meihui.wang.20@ucl.ac.uk}
\orcid{0000-0001-8420-2141}

\affiliation{%
  \institution{UCL SpaceTimeLab}
  \city{London}
  \country{UK}
}

\author{Stefano Cavazzi}
\email{stefano.cavazzi@os.uk}
\orcid{0000-0003-3575-0365}
\affiliation{%
  \institution{Ordnance Survey}
  \city{Southampton}
  \country{UK}
}

\author{James Haworth}
\email{j.haworth@ucl.ac.uk}
\orcid{0000-0001-9506-4266}

\affiliation{%
  \institution{UCL SpaceTimeLab}
  \city{London}
  \country{UK}
}


\begin{abstract}
The Common Crawl (CC) corpus is the largest open web crawl dataset
containing 9.5+ petabytes of data captured since 2008.
The dataset is instrumental in training large language models,
and as such it has been studied for (un)desirable content,
and distilled for smaller, domain-specific datasets.
However, to our knowledge, no research has been dedicated to
using CC as a source of annotated geospatial data.
In this paper, we introduce an efficient pipeline to extract annotated
user-generated tracks from GPX files found in CC,
and the resulting multimodal dataset with 1,416 pairings of human-written descriptions and \texttt{MultiLineString} vector data from the 6 most recent CC releases.
The dataset can be used to study people's outdoor activity patterns,
the way people talk about their outdoor experiences, as well as for developing trajectory generation or track annotation models, or for various other problems in place of synthetically generated routes. Our reproducible code is available on GitHub: \href{https://github.com/ilyankou/cc-gpx}{https://github.com/ilyankou/cc-gpx}.
\end{abstract}



\received{29 May 2024}

\maketitle

\section{Introduction}

The Common Crawl (CC) corpus\footnote{https://commoncrawl.org/} is the largest publicly available web crawl dataset containing 9.5+ petabytes of data dating back from 2008 \cite{baack_training_2024}. Due to its size, the dataset is widely used to train large language models, including GPT-3 \cite{thompson_whats_2022}.

In the past, researchers have studied the contents of Common Crawl for undesirable content \cite{luccioni_whats_2021}, extracted documents to create the largest Spanish-language crawling corpus \cite{gutierrez-fandino_escorpius_2022}, extracted parallel texts for machine translation \cite{smith_dirt_2013}, PDF documents \cite{turski_ccpdf_2023} and Word documents with layout annotations \cite{weber_wordscape_2023}, and built the `Colossal Clean Crawled Corpus' (C4) dataset\footnote{https://paperswithcode.com/dataset/c4} among other things. However, to our knowledge, no research was dedicated to analysing the geospatial contents or extracting annotated geospatial data from CC.

While websites and apps such as Strava\footnote{https://www.strava.com/}, AllTrails\footnote{https://www.alltrails.com/}, and Outdooractive\footnote{https://www.outdooractive.com/en/} contain large collections of curated and user-generated routes that are often accompanied by textual descriptions, these typically come with substantial licensing restrictions. CC, on the other hand, provides free access to all its data collected in accordance with website policies (subject to fair use restrictions), and exposes researchers to niche and varied websites that are harder to come across any other way.

In this paper, we introduce an efficient pipeline to identify, download, and clean GPX files that contain user-generated tracks, such as those produced by recording hiking, running, or cycling activities, along with human-written textual descriptions of those tracks. We run our process on the six most recent CC releases (spanning just over a year) to generate a multi-lingual, multi-activity dataset containing over 1,400 pairs of annotated \texttt{MultiLineString} features. When extended to all CC data back to 2008, the dataset is likely to contain over 10,000 samples.

The resulting dataset can be used to study the way people talk about outdoor activities, to facilitate the generation of human-like descriptions of activity tracks, or to study outdoor activity trajectories.

\section{Data Collection and Processing}

Hikers, cyclists, and in particular runners often track their activities using GPS-enabled mobile devices, such as Garmin smart watches \cite{karahanoglu_how_2021}, and sometimes add detailed annotations describing their experience. Alternatively, people may plan their active journey beforehand using websites such as \href{https://cycle.travel/}{\texttt{cycle.travel}}, and add notes before or after the journey. In this paper, we decided to focus on the popular XML-based \texttt{.gpx}\footnote{http://www.topografix.com/GPX/1/1/} (GPX Exchange Format) files that are most commonly used to share the activities, and ignore less popular formats such as \texttt{.fit} or \texttt{.tcx} which are very rare in Common Crawl.

\subsection{Downloading GPX files from CC}
We use index tables from the 6 most recent releases, \texttt{CC-MAIN-2023-*} and \texttt{CC-MAIN-2024-10}, to identify GPX files. Each index table contains a list of URLs and MIME-types of all web pages and files scraped by the CCBot in that release, and is broken into 300 parts. We use \texttt{duckdb}\footnote{https://duckdb.org/} to query each part to locate files whose detected MIME type is \texttt{ilike `\%gpx\%'} or whose filename extension is `.gpx' (typically both conditions are true at once).

For each GPX file, we record its Web Archive (WARC) file location, together with the offset and length in bytes. The offset and length indicate the location of the GPX file within a WARC file. This allows us to use Python's \texttt{requests}\footnote{https://github.com/psf/requests} library and the \textit{Range} request header parameter to download individual GPX files from CC without the need to download large WARC files first:

\begin{verbatim}
requests.get(
    f'https://data.commoncrawl.org/{warc_file}'
    headers={
        'Range': f'bytes={offset}-{offset+length-1}'
    })
\end{verbatim}

This massively speeds up the data acquisition process; in fact, it takes approximately 40 minutes to download all GPX files from a single CC release, which typically contains around 90 terabytes of compressed data spread across 3 billion files. We identified 112,953 GPX files across the six most recent CC releases. We were able to successfully download 111,102 files (98.4\% of those identified). Among downloaded files, 102,103 (or 91.9\%) came from unique URLs, indicating that very few GPX files appear in more than one CC release. After removing duplicate GPX files that come from different URLs, we are left with 94,170 GPX files.

\subsection{Identifying high-quality tracks}

We use \texttt{gpxpy}\footnote{https://github.com/tkrajina/gpxpy} Python library to parse response strings into GPX objects that can be analysed. A typical GPX file consists of tracks; tracks consist of segments; segments consist of points. GPX files representing simple recorded point-to-point activities, such as hiking or cycling trips, would normally contain a single track that consists of one or more segments (for example, a runner may choose to record each lap as a separate segment). Multi-track GPX files that we inspected were typically used to record multi-day activities such as races around Europe. We decided to keep only single-track GPX files that represent activities no longer than 100 km (62.1 mi) because we believe longer activities cannot be adequately described in a few paragraphs, and are unlikely to be relevant to most people. We also remove activities shorter than 0.5 km (0.31 mi).

We keep tracks that have at least 1 GPS point per 100 m (328 ft) on average ($\sim$87\% of all tracks satisfy this constraint). The threshold was determined empirically by analysing individual GPX files.

\subsection{Identifying activity descriptions}

We extract descriptions from each track's \texttt{<desc>} tag. We use \texttt{BeautifulSoup}\footnote{https://www.crummy.com/software/BeautifulSoup/} Python library to remove occasional HTML tags present in text, replace characters such as newlines and tabs with single spaces, and use regular expressions to remove text inside square and curly brackets (tags likely added by some apps).

We only keep the tracks that have an associated description between 50 and 2000 characters long after the manipulations above. The upper character limit represents $\sim$$99^{th}$ percentile of the description length distribution.

To identify high-quality track and activity descriptions, we employ Llama-3 8B Instruct, one of the most capable open-source language models as of April 2024 \cite{meta_ai_introducing_2024}. We use the prompt \textit{Does the text in triple quotes represent a high-quality and insightful route or track description, or an activity description such as hiking, cycling, or racing? Respond with `True' or `False'. If you are unsure, say `False'. Text: ```\texttt{\{text\}}'''}, and  remove tracks for which \textit{False} is returned.

Positive examples (tracks included in the dataset):

\begin{itemize}
    \item `A lovely ~4 hour walk from Ockley railway station to Dorking, via the beautiful view from Leith Hill Tower.'
    \item `Quieter roads and backstreets, quirky interest but still direct.'
    \item `This trail leading to the borough of Kukleny takes in significant industrial buildings evoking the glory of the iron and leather industries, as well as family houses, a funeral hall, and a Cubist vocational school.'
\end{itemize}

Negative examples (tracks excluded from the dataset):

\begin{itemize}
    \item `Cheese-making is a centuries-old tradition in Gruyère and firmly rooted in the heritage. Immerse yourself in a walk with a historic feel.'
    \item `File with points/tracks from Locus Map Classic/3.65.2'
    \item `Note check opening times of Manor Park Cremitorium'
\end{itemize}

\subsection{Removing personal information}

The GPX files available in CC would typically be uploaded and openly shared by the users in the open web. However, it is our responsibility as researchers to respect user privacy and remove any information that can be deemed personal.

We use regular expressions to mask emails and URLs (very occasionally found in descriptions) with tokens \textit{<EMAIL>} and \textit{<URL>}. We also identify phone numbers using Google's \texttt{libphonenumber}\footnote{https://github.com/google/libphonenumber} Python library, and mask those with \textit{<TELEPHONE>}. We also remove timestamps from points due to both privacy and data quality reasons (many tracks lack timestamps, and, unlike missing elevation data, timestamps are impossible to recreate).

We then use Llama-3 8B Instruct with the prompt \textit{Does the text in triple quotes contain any personally identifiable information, such as someone's address or name? Respond with `True' or `False'. If you are unsure, say `True'. Text: ```\texttt{\{text\}}'''} to further identify and remove tracks whose descriptions may contain personal information.

\subsection{Translating descriptions to English}

The majority of GPX files we identified are from European websites, and their descriptions are not in English. We use \texttt{pycld2}\footnote{https://github.com/aboSamoor/pycld2} Python library to identify the original language. We remove tracks whose language is identified as \textit{Unknown} as those often contain majority numbers or incoherent text, as well as all tracks whose descriptions are in languages that have 5 or fewer samples.

Initially, we experimented with several popular open-source LLMs, including Mistral and the Llamas (2 \& 3), to translate descriptions into English, a task LLMs are shown to be good at \cite{zhu_multilingual_2023}. Unfortunately, we discovered that LLMs occasionally modify formatting and add unnecessary context (e.g., \textit{`The text is in Ukrainian. Here's the English translation: ...'}). Extensive prompt-engineering was not able to remedy these issues. We ultimately chose to use \texttt{argos-translate}\footnote{https://github.com/argosopentech/argos-translate}, a popular offline Python translation library based on \texttt{OpenNMT}\footnote{https://opennmt.net/}, an open source neural machine translation system, to translate all non-English descriptions into English.

\subsection{Adding missing elevation}

Just over half of all relevant GPX files contained device-recorded point elevation data. To produce a consistent dataset with all tracks represented by 3D points (lat, lon, elevation), we use the Shuttle Radar Topography Mission\footnote{https://www.earthdata.nasa.gov/sensors/srtm} digital elevation model, acquired via the \texttt{SRTM.py}\footnote{https://github.com/tkrajina/srtm.py} Python library, to approximate point elevations in tracks where elevation data is not originally available. We note the source of the elevation data, \textit{GPS} (device-recorded) or \textit{DEM} (SRTM), in the \textit{elev\_source} field (see Table \ref{tab:dataset_properties}).

\section{Dataset \& Potential Applications}

\begin{figure}[ht]
  \centering
  \includegraphics[width=\linewidth]{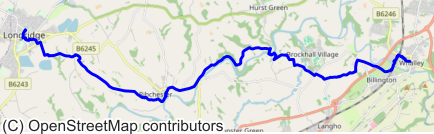}
  \caption{A 18.2 km (11.3 mi) route in the UK. The description reads: \textit{`Longbridge to Whalley Slowway following part of the Ribble Way. Difficult to find a good crossing of the A59. The crossing chosen crosses the road from footpath to footpath in a place with good visibility. The road junctions/bridges were actually worse as would need to walk along a fast road with no pavement rather than just cross once at right angles. This crossing sets up good sections without roads. Good spacing of waypoints at Old Langho and Ribchester.'}}
  \Description{Example of a linear route from the dataset.}
  \label{gpx_uk}
\end{figure}

\begin{figure}[ht]
  \centering
  \includegraphics[width=\linewidth]{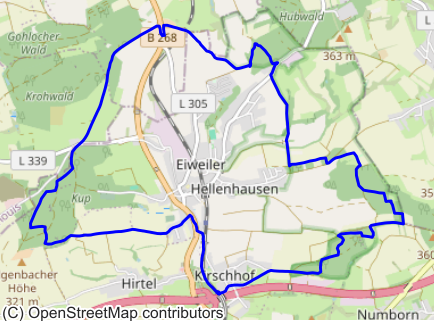}
  \caption{A 13.8 km (8.6 mi) circular route in Germany. The description in German reads: \textit{`Der Weg ist sehr gut gekennzeichnet mit einem schwarzen Hirschkäfer (Hootzemann) auf weißem Grund. Mein Start- und Zielpunkt war das Schützenhaus Eiweiler in der Nähe der Großwald-Brauerei.'} The English translation is: \textit{`The path is very well marked with a black deer beetle (Hootzemann) on white ground. My starting and finishing point was the Schützenhaus Eiweiler near the Großwald brewery.'}}
  \Description{Example of a circular route from the dataset.}
  \label{gpx_germany}
\end{figure}

The final dataset consists of 1,416 tracks (or around 1.5\% of the initially deduplicated GPX files) and accompanying descriptions in both the original language and English, along with other properties described in Table~\ref{tab:dataset_properties}. The examples of two tracks with descriptions are shown in Figures~\ref{gpx_uk} and \ref{gpx_germany}.

\begin{figure*}[ht]
  \centering
  \includegraphics[width=\linewidth]{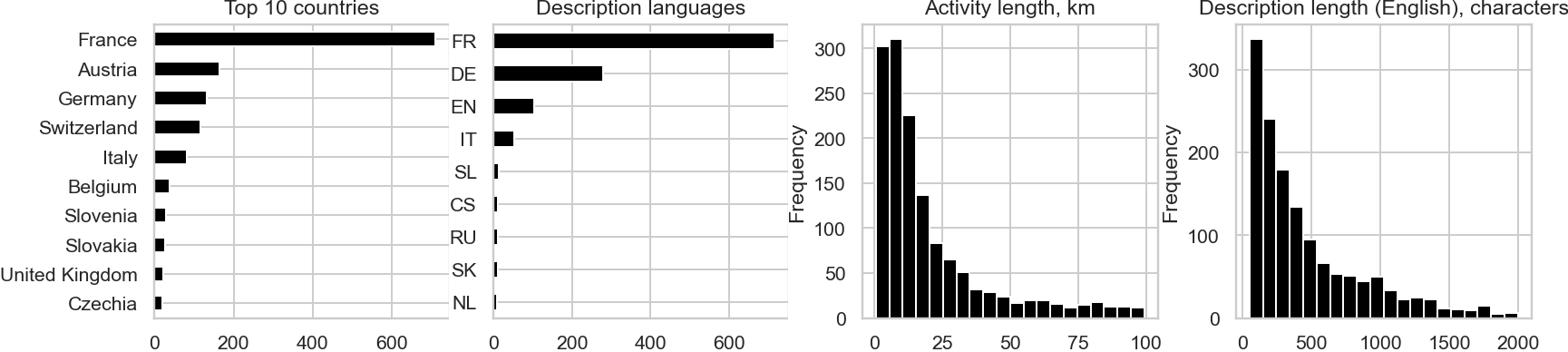}
  \caption{Select dataset properties.}
  \Description{Select dataset properties.}
  \label{gpx_stats}
\end{figure*}

The GPX files come from 135 unique domain names. Track descriptions come in 11 languages, with the majority being French (802), German (330), and English (110). The original descriptions are between 50 and 1999 characters long (for translated, between 34 and 1866), with a median of 303 characters (for translated, 285.5). The tracks originate in 25 countries, with most popular being France (710), Austria (164), Germany (133), Switzerland (116), and Italy (81). The shortest track is 607 m (1991 ft) while the longest is 99.3 km (61.7 mi). The average track length is 20.5 km (12.7 mi), while the median is much shorter at 12.1 km (7.5 mi). Histograms of route length (in km) and description length (in characters) are shown in Figure~\ref{gpx_stats}.

\begin{table}
  \caption{Dataset property descriptions}
  \label{tab:dataset_properties}
  \begin{tabular}{rl|p{5.5cm}}
    \toprule
    \# & Property & Description\\
    \midrule
    1 & url & URL of the GPX file\\
    2 & warc\_file & CC WARC file with GPX file \\
    3 & warc\_offset & GPX file position in WARC \\
    4 & warc\_len & GPX file byte length \\
    5 & country & Country name as determined by the first point in the track intersecting boundaries from https://www.geoboundaries.org/ \\
    6 & desc & Original track description \\
    7 & desc\_lang & Track description language code, as determined by \texttt{pycld2} \\
    8 & desc\_en & Track description translated into English \\
    9 & elev\_source & \textit{GPS} if elevation is recorded by device; \textit{DEM} if determined later from Shuttle Radar Topography Mission \\
    10 & elev\_highest & Track's highest point, m \\
    11 & elev\_lowest & Track's lowest point, m \\
    12 & uphill & Cumulative elevation gain, m \\
    13 & downhill & Cumulative elevation loss, m \\
    14 & length\_2d & Track length disregarding elevation, m \\
    15 & length\_3d & Track length accounting for elevation, m \\
    16 & is\_circular & \textit{True} if start and end points are within 350 m from each other, \textit{False} otherwise\\
    17 & geometry & MultiLineString Z geometry in GPS coordinates: \textit{(lat, lon, elevation)} \\
  \bottomrule
\end{tabular}
\end{table}

The tracks in the resulting dataset represent real-life outdoor activities that were either recorded (i.e., completed) or planned using GIS software. Thus, these routes can be used in place of synthetically generated trajectory datasets, as well as for training trajectory generation ML models that are focused on recreational activities.

The description-\texttt{MultiLineString} pairs can be used to study how people describe their outdoor activities, such as what kinds of landmarks, landscapes, and experiences they choose to mention or ignore, and get an insight into turn-by-turn decisions of users. These pairs can also be used to fine-tune language models to generate human-like descriptions of routes, or to compute location embeddings. Our dataset can become especially powerful when combined with Point of Interest (POI) and/or road segment datasets, such as those from OpenStreetMap\footnote{https://welcome.openstreetmap.org/} or Overture\footnote{https://overturemaps.org/}.

\section{Conclusion}

In this paper, we demonstrated that Common Crawl can be used as a novel source of annotated geospatial data. As a case study, we built an efficient pipeline to extract annotated outdoor activity tracks from GPX files that can be executed on a medium-powered laptop in a matter of hours. The resulting multimodal dataset contains textual descriptions of outdoor activities paired with \texttt{MultiLineString} vector geospatial data and can be used to study people's outdoor activity habits, and the relationship between what people experience and what they \textit{say} they experience. The dataset can be useful as a fine-tuning set for various GeoAI models, from trajectory to human-like description generators. We recognise that our case study captures just a fraction of all annotated geospatial data available in CC, and we intend to extend the scope in future work.

\begin{acks}
This work was supported by Ordnance Survey \& UKRI Engineering and Physical Sciences Research Council [grant no. EP/Y528651/1].
\end{acks}

\bibliographystyle{ACM-Reference-Format}
\bibliography{articles}


\end{document}